\documentclass[runningheads]{llncs}
\usepackage[T1]{fontenc}
\usepackage{graphicx}
\usepackage{booktabs}
\usepackage[misc]{ifsym}
\newcommand{\corr}{(\Letter)}

\begin{document}

\title{From Sea to System: Exploring User-Centered Explainable AI for Maritime Decision Support}

\titlerunning{From Sea to System}

\author{Doreen Jirak\inst{1} \corr \orcidID{0009-0003-2839-3475} \and
Pieter Maes\inst{2}  \and Armees Saroukanoff \inst{2} \and Dirk van Rooy \inst{1}}

\authorrunning{D. Jirak et al.}

\institute{University of Antwerp, Paardenmarkt 94, 2000 Antwerp, Belgium \email{\{doreen.jirak,dirk.vanrooy\}@uantwerpen.be}
\and
Antwerp Maritime Academy (AMA), Noordkasteel Oost 6, 2030 Antwerp, Belgium \email{\{pieter.maes, armeen.saroukanoff\}@hzs.be}
}

\maketitle              

\begin{abstract}
As autonomous technologies increasingly shape maritime operations, understanding why an AI system makes a decision becomes as crucial as what it decides. In complex and dynamic maritime environments, trust in AI depends not only on performance but also on transparency and interpretability. This paper highlights the importance of Explainable AI (xAI) as a foundation for effective human-machine teaming in the maritime domain, where informed oversight and shared understanding are essential. To support the user-centered integration of xAI, we propose a domain-specific survey designed to capture maritime professionals’ perceptions of trust, usability, and explainability. Our aim is to foster awareness and guide the development of user-centric xAI systems tailored to the needs of seafarers and maritime teams.

\end{abstract}

\section{Introduction}
\label{sec:intro}
The maritime industry is undergoing a profound transformation. With advances in artificial intelligence, sensor fusion, and remote operation, Maritime Autonomous Surface Ships (MASS) \cite{Li19} are no longer speculative concepts but emerging realities. Maritime tasks such as remote bridge operations \cite{Bhuiy24} call to redefine the roles and responsibilities of seafarers and maritime authorities alike. Yet, this shift is taking place in a domain known for its conservatism. Unlike the aviation sector, which has a long history of automation and centralized regulation, the maritime sector is highly decentralized and slow to adapt. Operational decisions at sea are shaped not only by technical constraints but also by weather conditions, team dynamics, and longstanding seafaring practices. As such, the integration of autonomous and hybrid systems into maritime operations presents a unique set of challenges.
Explainable AI (xAI) offers a promising avenue to address the trust gap between maritime professionals and AI systems \cite{Veitc21}. By making the rationale behind AI-driven decisions transparent and interpretable \cite{Merwe23}\cite{Madse24}, xAI has the potential to foster user confidence and support safe collaboration between human operators and intelligent systems. In the context of daily maritime routines where adherence to the COLREGs (International Regulations for Preventing Collisions at Sea) is essential, xAI could help clarify how autonomous systems interpret these rules and adapt in real time.

This paper presents the design of an empirical study aimed at evaluating maritime stakeholders’ attitudes toward explainable AI in navigational decision support systems. We aim to investigate how explainability features affect user trust, perceived usefulness, and willingness to engage with AI-assisted systems. To this end, we developed a survey-based experimental framework that includes scenario-based stimuli, pre and post-questionnaires assessing trust, openness to technology, and user satisfaction, as well as interactive elements grounded in realistic radar-based tasks. Given the ongoing nature of data collection and the potential for participant bias, this paper focuses on the conceptual framework and design rationale rather than providing detailed stimuli or full survey items. By presenting a survey framework grounded in real-world navigational scenarios, this work aims to inform the development of explainable, trustworthy AI systems tailored for the maritime domain. The results will provide early insight into human-AI interaction challenges and guide future research in the design of interpretable, regulation-compliant autonomous systems at sea. This way, we hope to create more awareness not only the human factors involved in (team) decision-making, but also to foster interdisciplinary research towards \textit{user-centered} xAI.


\section{Related Work}
\label{sec:rw}
The maritime sector, while historically conservative in adopting automation, is now facing a wave of technological transformation driven by advances in artificial intelligence (AI). Applications range from decision-support in navigation tasks to fully autonomous control aboard Maritime Autonomous Surface Ships (MASS) \cite{Li19}. These developments promise increased operational efficiency and economic benefits, yet they also raise critical questions about how such technologies can be reasonably integrated into the complex and uncertain environment of maritime operations. For instance, a major concern remains maritime safety. Despite technological progress, human error continues to be a dominant cause of accidents at sea.
A commonly cited number stating that around 80\% of maritime accidents are due to human error dates back to a study released by Berg et al. \cite{Berg13}. However, depending on the time range and the territorial waters under investigation, these numbers fluctuate around 60\% \cite{Erol15} and up to 96\% \cite{Sanch21}. Root causes often include cognitive fatigue, miscommunication, situational overload, or mental blocking, particularly under pressure or in dynamic team settings. These findings underscore the need to not only automate tasks, but also to complement human performance and enhance decision-making reliability in safety-critical scenarios.
Therefore, the integration of AI systems in maritime contexts must go beyond technical robustness. We argue that it must be \textit{user-centered}, recognizing the operational routines, domain knowledge, and lived experiences of seafarers. This calls for a holistic, interdisciplinary approach that considers both human and machine capabilities. In particular, insights from human factors research, cognitive science, neuroscience, and social psychology are essential to understand how seafarers interpret, adapt to, and interact with these emerging technologies. 
A central prerequisite for the transition to user-centered human-AI interaction and teaming is trust \cite{Mayer95}. Trust in automation is not a monolithic construct but encompasses trust calibration, both cognitive trust (e.g., perceived competence, reliability, and predictability of the system) and affective trust (e.g., feelings of safety, alignment with human intentions, and emotional comfort with the system’s behavior) \cite{Legoo23}, and trust development over time. In high-risk environments like the maritime domain, where accidents can have serious consequences, designing to enhance trust in AI systems is essential. This leads to the important aspect of how to create also trustworthy systems and interactions in the maritime sector. Unlike controlled simulations, real-world trust is developed and shaped by experience, institutional trainings, and team practices on the bridge. It is not only a matter of whether an AI system or autonomous agent makes ``correct” decisions, but whether its decisions are understandable, predictable, and aligned with user expectations \cite{Hoffm23}, especially when system actions diverge from standard maritime practices.

In this context, explainable AI (xAI) is a key enabler of trustworthy and effective human-AI collaboration. As deep learning models (often simply ``AI”) have demonstrated superhuman performance in detection and recognition tasks, their lack of decision transparency has raised concerns. xAI addresses this by developing methods to open the, often called, ``black box” of AI decision-making, enhancing transparency, interpretability, and fairness.
While recent user-centered xAI has focused on increasing user trust \cite{Liao22} and preventing over-reliance \cite{Bucin21}, the maritime sector has received comparatively little attention in this regard. However, a recent study by Merwe et al. \cite{Merwe24}
showed that agent transparency can potentially improve situational awareness (SAW \cite{Endsl23}) in seafarers but might be a function of traffic complexity and richness of the agent's reasoning. Similarly, Madsen et al. \cite{Madse25} revealed that agent transparency did not improve all SAW levels uniformly and that information display needs calibration to the human cognitive resources, fostering further the need for user-centered xAI studies. To address this gap, we propose a survey-based study aimed at capturing how maritime professionals perceive explainability, trust, and usability in AI-driven decision-making systems. Our goal is to lay the groundwork for \textit{user-centered} xAI interaction design tailored to the unique cognitive and operational realities of the maritime domain.

\section{Survey Design}
\label{sec:design}
As part of our ongoing research, we propose a survey aimed at capturing user perceptions of trust, usability, and explainability in the context of maritime autonomy. Our goal is to raise awareness, map trust dimensions (both cognitive and affective), and identify design priorities for future human-centered xAI systems. This effort contributes not only to MASS development but also to the broader integration of explainable and trustworthy AI in safety-critical, operationally demanding domains.
Our research is guided by the following main research questions:

\begin{enumerate}
    \item RQ1: What is the seafarer's disposition to technological progress and their propensity to trust? (pre-questionnaire)
    \item RQ2: To what extent does the inclusion of explainability features (xAI, ``maritime assistant") affect user satisfaction, trust, and eventually willingness to adopt such systems? (post-questionnaire)
    \item RQ3: How do maritime professionals perceive the usefulness and trustworthiness of the maritime assistant? What are the concerns and expectations of seafarers regarding the integration of AI into maritime daily routines and team workflows? (post-questionnaire, open questions)
\end{enumerate}

\begin{figure}[htbp]
    \centering
    \includegraphics[width=0.8\textwidth]{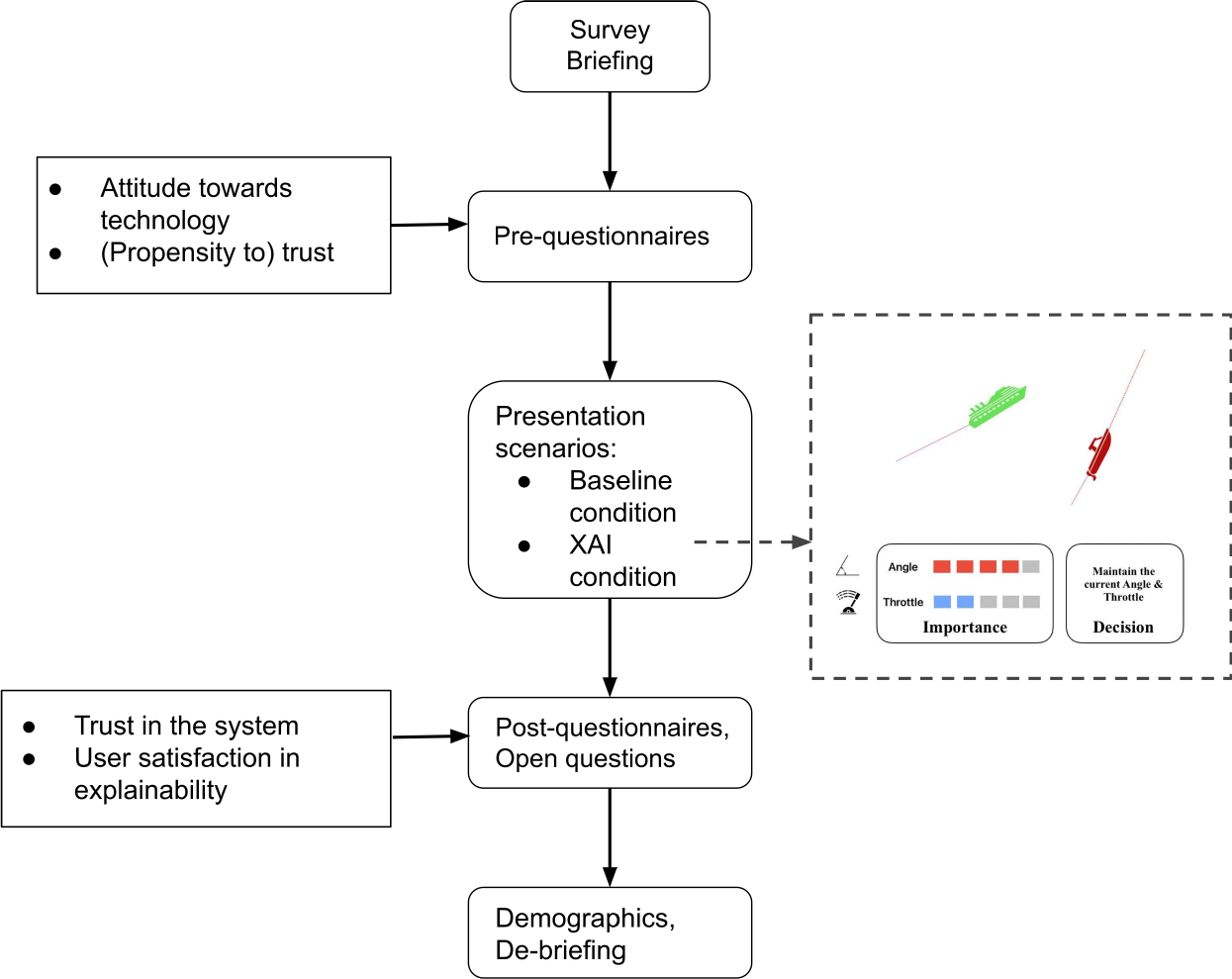}
    \caption{Experimental flowchart of the prospective maritime xAI survey.}
    \label{fig:flow}
\end{figure}

Figure \ref{fig:flow} shows the experimental flow of the survey, structured into:
\begin{enumerate}
    \item Briefing: Welcome and explanation of the upcoming presentation. The participant gives their consent (GDPR) and is granted drop out of the survey at any moment and without facing any disadvantages. Further, no data will be collected from participants who wish to drop out.
    \item Survey: The concrete survey contains some pre-questionnaires, followed by the presentation of scenarios drawn from daily maritime routines on a vessel, e.g., collision avoidance. Information is given by radar images which are selected based on critical time steps, for instance, when COLREG rules do not apply. After obtaining the sailor's responses about their action suggestion (baseline condition), we present the output of a ``maritime assistant" (cf. Figure \ref{fig:flow}) displaying its decision and the most significant feature involved in the decision-making (xAI condition). After the presentation of the scenarios, post-
    questionnaires apply checking for the user satisfaction and trust in autonomous systems. As described, explainability must go hand in hand with usability. The design of user-centered xAI in this context should therefore consider factors beyond system performance, including how well the explanations support timely, safe, and confident human decisions. While standard scales for system usability do not apply due to absence of real interactions, still xAI metrics like suggested in \cite{Hoffm23} apply and can be merged together with individual, scenario-tailored items. Additionally, participants can give their feedback to the presented scenarios.
    \item De-briefing: Collection of demographic data including the participant's rank and years of sea experience. We decided to ask for this data after the survey to avoid any biases introduced by age or experience-related questions. The survey closes with the question of database inclusion for further experiments and explanation of the survey objectives.
\end{enumerate}

Once data collection is complete, future work will report on findings from this study and provide a validated survey instrument for broader use. We also anticipate expanding this research to explore how explainability preferences vary across roles (e.g., navigators vs. engineers) and levels of AI familiarity.
\section{Conclusion}
In this paper, we argue that as we introduce intelligent systems into the maritime domain, we must move beyond purely technical narratives but need to include human factors in the decision-making process such as cognitive resources and affective states, but also trust and willingness to technology adoption. Therefore, we introduce our ongoing research on user-centered xAI incorporating these human factors coupled with important domain knowledge from prospective user in maritime human-AI decision-making.  
We hope to foster awareness among seafarers about the upcoming technological changes but also to convince (x)AI developers to build \textit{user-centric} interfaces to create meaningful AI integration, especially in dynamically interacting teams such as maritime crew members. To succeed, hybrid learning and decision-making systems must align not only with performance metrics but also with the lived experience and expertise of those at sea.

\begin{credits}
\subsubsection{\ackname} The authors thank Arian Sabaghi Khameneh (imec/IDLab) for the provision of the xAI image. This work is conducted within the DEFRA AHOI project, funded by the Belgian Royal Higher Institute for Defence, under contract number 23DEFRA002.

\subsubsection{\discintname}
The authors have no competing interests to declare that are
relevant to the content of this article.
\end{credits}
%
%
%
\bibliographystyle{splncs04}
\bibliography{Jirak_FromSeaToSystem_2025}
%




\end{document}